\documentclass[11pt]{article}

\usepackage[margin=1in]{geometry}
\usepackage{lmodern}
\usepackage[T1]{fontenc}
\usepackage{graphicx}
\usepackage{subcaption}
\usepackage{booktabs}
\usepackage{amsmath,amssymb}
\usepackage{xcolor}
\usepackage[numbers,sort&compress]{natbib}
\usepackage{float}
\usepackage{multirow}
\usepackage{array}
\usepackage{enumitem}
\usepackage{caption}
\usepackage[colorlinks=true, citecolor=blue, linkcolor=blue, urlcolor=blue]{hyperref}

\title{\textbf{Brain-to-Image Retrieval and Reconstruction\\ via Multimodal EEG Alignment}\\[0.5em]
\large }

\author{
  \normalfont Chi Kit Wong \\ ckwong627@connect.hkust-gz.edu.cn\\
  \and
  Yan Liu \\
  yliu674@connect.hkust-gz.edu.cn\\
  \and
  Haowen Yan \\ hyan904@connect.hkust-gz.edu.cn\\
}

\date{}

\begin{document}

\maketitle

\begin{abstract}
We present a brain-to-image system that decodes visual stimuli from EEG signals recorded during natural image viewing.
Our system addresses two mandatory tasks: (1) \emph{EEG-to-Image Retrieval}, which ranks the correct stimulus image among 200 candidates given an EEG segment, and (2) \emph{EEG-to-Image Reconstruction}, which generates an image consistent with the perceived stimulus.
For retrieval (Task~1), we implemented an approach of multi-level blurring~\citep{ved2026}, improved with biologically-inspired EVNet~\citep{evnet2025} features, trained with the InfoNCE loss~\citep{oord2018cpc}.
Evaluated over 10 random seeds, we achieve a mean final-epoch Top-1 accuracy of \textbf{86.30\%} and Top-5 accuracy of \textbf{98.55\%} for a single subject.
For reconstruction (Task~2), we implemented CognitionCapturerPro (CogCapPro)~\citep{ccp2026} that aligns EEG representations to multi-modal CLIP embeddings (image, text, depth, edge) and synthesizes images with SDXL-Turbo~\citep{sauer2023sdxlturbo} conditioned via IP-Adapter~\citep{ye2023ipadapter}.
Averaged over 10 seeds, our system achieves a CLIP Score of \textbf{0.903} (ViT-H-14, paper-comparable) / \textbf{0.870} (ViT-L/14, course-standard) and SSIM of \textbf{0.409}. The code is available in \url{https://github.com/Chikit-WONG/DL_Project/}
\end{abstract}

\section{Introduction and Problem Formulation}

Understanding how the human brain encodes visual information is a central challenge in computational neuroscience and brain-computer interfaces.
Recent advances in deep learning have made it possible to decode EEG signals recorded during image viewing into meaningful visual representations.
In this project, we tackle two complementary tasks using a preprocessed version of the THINGS-EEG2 dataset~\citep{gifford2022thingseeg2data}. For retrieval, we build on a multi-blur visual representation and augment it with an EVNet~\citep{evnet2025} front end; for reconstruction, we adapt CognitionCapturerPro (CogCapPro)~\citep{ccp2026}.

\paragraph{Dataset.}
The provided preprocessed THINGS-EEG2 files contain repeated EEG responses for each stimulus.
We use preprocessed EEG data from a single subject with 63 channels sampled at 250\,Hz.
Each response is represented by a 250-sample post-stimulus segment, corresponding to approximately 1 second.
For both training and testing, we average over the repetition dimension of the provided tensors, yielding 16,540 training EEG-image pairs and 200 held-out test samples under the official 200-way protocol.

\paragraph{Task 1: EEG-to-Image Retrieval.}
Given an EEG segment $\mathbf{e} \in \mathbb{R}^{63 \times 250}$, rank a set of 200 candidate images such that the ground-truth stimulus image appears as high as possible.
Performance is measured by Top-1 and Top-5 accuracy (the rate the correct image class appeared as the first ranking or in the first-five rankings).

\paragraph{Task 2: EEG-to-Image Reconstruction.}
Given an EEG segment $\mathbf{e}$, generate an image that is semantically and perceptually consistent with the viewed stimulus.
Performance is measured by CLIP Score~\citep{radford2021clip}, feature-based reconstruction metrics computed by the provided evaluation script: SSIM~\citep{wang2004ssim}, AlexNet-~\citep{alexnet2012} and Inception-based two-way identification scores, SwAV~\citep{swav2021}, and EfficientNet~\citep{efficientnet2020} correlation distance.

\section{Method}

\subsection{Task 1: Dual-Stream Contrastive Retrieval}

\paragraph{EEG Encoder.}
The EEG signal $\mathbf{e} \in \mathbb{R}^{63 \times 250}$ is processed by an encoder~\citep{ved2026}:
\begin{enumerate}[leftmargin=*, topsep=2pt, itemsep=1pt]
  \item A 2D convolution with absolute-value activation ($1 \to 25$ channels, kernel size $63 \times 1$), producing $25$ feature maps of dimension $250$.
  \item Batch normalization.
  \item A shared MLP applied independently to each of the $25$ feature maps: linear $250 \to 200$ with ELU and dropout ($p = 0.25$), followed by linear $200 \to 200$ with ELU and dropout ($p = 0.65$).
  \item Flattening to a $5000$-dimensional vector.
  \item A final linear projection to a shared embedding of dimension $1024$.
\end{enumerate}
The EEG encoder outputs a 1024-dimensional embedding $\mathbf{z}_\text{EEG} \in \mathbb{R}^{1024}$.

\paragraph{Multi-Blur Visual Stream.}
Rather than using a single image representation or the 10 blurring levels used in prior work~\citep{ved2026}, we apply 8 levels of Gaussian blur under convolution kernel size $\in \{1, 3, 15, 21, 33, 45, 57, 63\}$ pixels to each training image and extract features from each blurred version using a frozen CLIP RN50 encoder~\citep{radford2021clip} (as also studied in~\citep{cherti2023openclip}).
The 8 feature vectors are first aggregated via a learned attention mechanism (softmax-weighted sum) into a single 1024-dimensional blur feature.
This multi-scale representation mimics the progressive spatial acuity of human visual processing.

\paragraph{EVNet Biologically-Inspired Stream.}
In parallel, we pass each image through the front end of EVNet pipeline~\citep{evnet2025}, which consists of a subcortical preprocessing block followed by a primary visual cortex (V1) simulation block (VOneBlock~\citep{vone2020}).
The EVNet output is subsequently processed by the same frozen CLIP RN50 backbone, yielding a complementary 1024-dimensional feature.

\paragraph{Stream Fusion.}
When EVNet is enabled, the aggregated blur feature and the EVNet feature are combined via a learned softmax gate:
\begin{equation}
  \mathbf{v}_\text{fused} = w_1 \cdot \mathbf{v}_\text{blur} + w_2 \cdot \mathbf{v}_\text{EVNet}, \quad (w_1, w_2) = \text{softmax}(\boldsymbol{\alpha})
\end{equation}
where $\boldsymbol{\alpha} \in \mathbb{R}^2$ is a learnable parameter.
The fused feature is then passed through a shared two-layer fusion adapter (1024 $\to$ 768 $\to$ 1024, dropout $=0.85$), producing the final image embedding $\mathbf{v} \in \mathbb{R}^{1024}$.
If EVNet is disabled, the aggregated blur feature is passed through the corresponding blur-only adapter instead.

\paragraph{Contrastive Training.}
We minimise the symmetric InfoNCE loss~\citep{oord2018cpc,chen2020simclr} over EEG--image pairs within each mini-batch:
\begin{equation}
  \mathcal{L} = -\frac{1}{2N}\sum_{i=1}^N \left[\log\frac{e^{\mathbf{z}_i \cdot \mathbf{v}_i}}{\sum_j e^{\mathbf{z}_i \cdot \mathbf{v}_j}} + \log\frac{e^{\mathbf{v}_i \cdot \mathbf{z}_i}}{\sum_j e^{\mathbf{v}_i \cdot \mathbf{z}_j}}\right]
\end{equation} 
In our implementation, the logits use a fixed scale of $1$ rather than a learned temperature.
At retrieval time, both EEG and visual embeddings are $\ell_2$-normalized before cosine-similarity ranking over the 200 candidate images.


\begin{figure}
  \centering
  \includegraphics[width=1.0\linewidth]{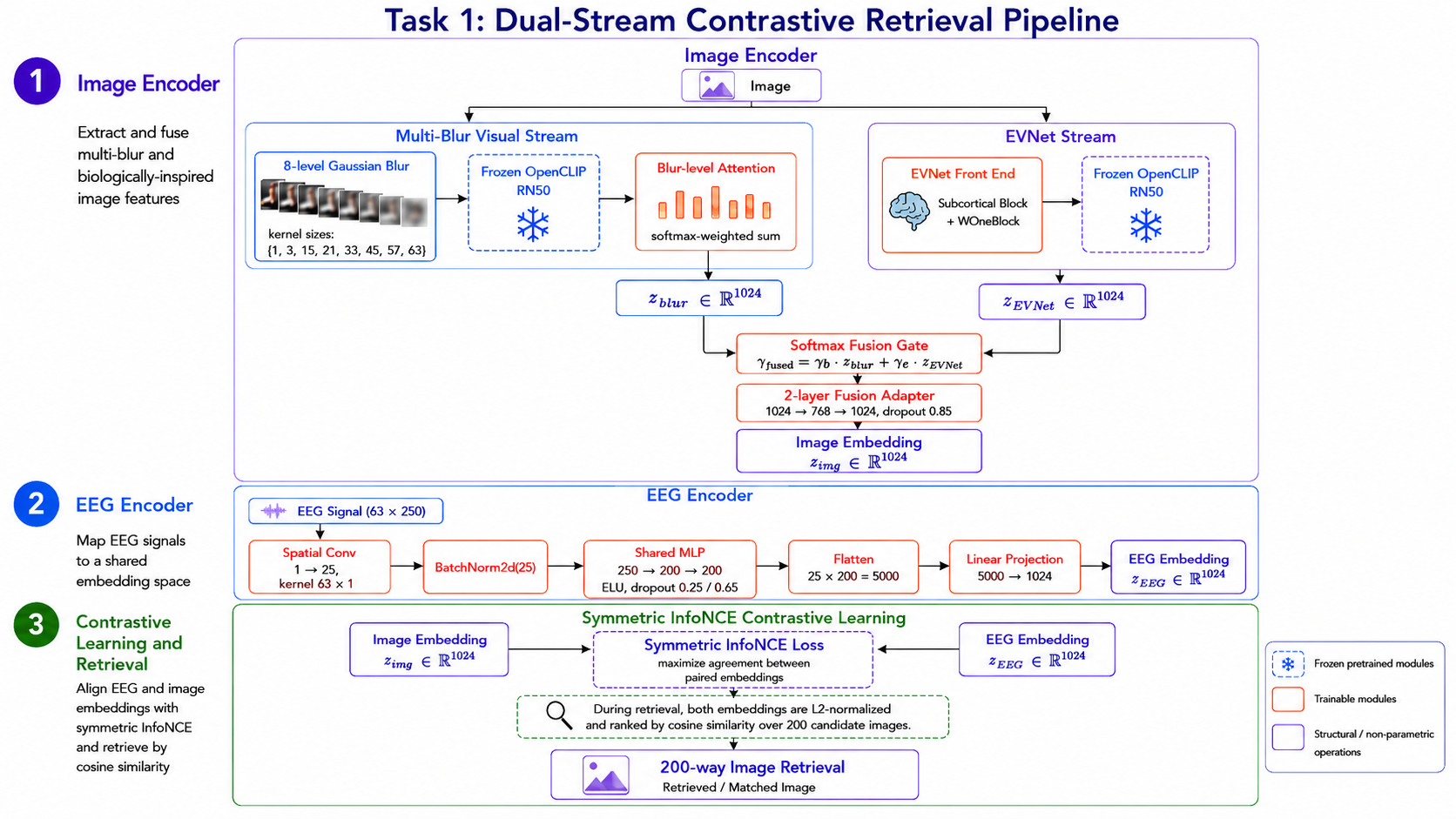}
  \caption{Image Retrieval Training Pipeline.}
  \label{fig:task1_pipeline}
\end{figure}

\subsection{Task 2: Multi-Modal EEG-to-Image Reconstruction}


\begin{figure}
    \centering
    \includegraphics[width=0.9\linewidth]{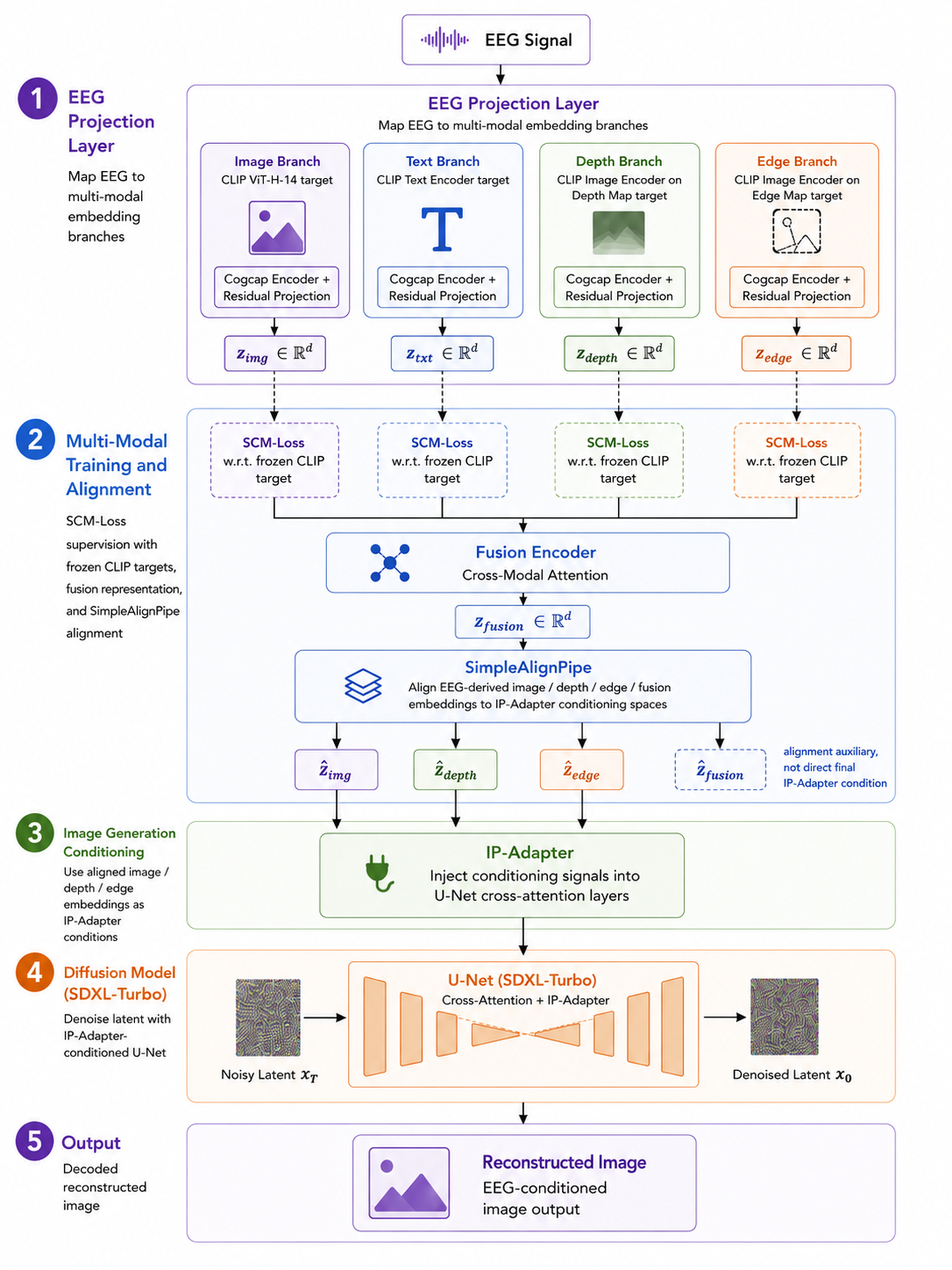}
    \caption{Pipeline of the Training Process of Image Reconstruction.}
    \label{fig:task2_pipeline}
\end{figure}

Task~2 is implemented using the CognitionCapturerPro (CogCapPro)~\citep{ccp2026} framework, which maps EEG signals to multiple CLIP-aligned embedding spaces and then drives a diffusion model to synthesise images.

\paragraph{EEG Projection Layer.}
A dedicated \texttt{EEGProjectLayer} uses four parallel CogCapPro branches to map the same EEG signal to four embedding spaces corresponding to four modalities: \emph{image} (CLIP ViT-H-14), \emph{text} (CLIP ViT-H-14 text encoder), \emph{depth map}, and \emph{edge map}. For depth and edge modalities, we first render depth maps from the stimulus images and apply edge detection, then extract their CLIP image features using the frozen CLIP ViT-H-14 image encoder. The four heads play complementary but not identical roles in reconstruction. The image head provides the most direct visual-semantic supervision, as it is trained to predict the CLIP image embedding extracted from the original stimulus image. This target embedding preserves category-level appearance, global shape, and colour information, making the image head the primary source of visual content for generation. The depth head is trained against CLIP image embeddings extracted from rendered depth maps, so it provides spatial and geometric layout cues. The edge head is trained against CLIP image embeddings extracted from edge maps, allowing it to impose contour and shape constraints. In contrast, the text head is trained against CLIP text embeddings and mainly acts as auxiliary semantic supervision, contributing indirectly through the shared EEG representation and fusion pathway. Each EEG modality branch is implemented as an independent encoder, followed by a residual projection module that produces an EEG-side latent vector $\mathbf{z}_m \in \mathbb{R}^{d}$ for modality $m$.

\paragraph{Multi-Modal Alignment.}
Each single-modality branch is trained with the Similarity-Category Masked Loss (SCM-Loss) described in CogCapPro~\citep{ccp2026}. Following the naming convention in the official CogCapPro implementation, our code implements this objective as \texttt{ClipLoss\_Modified\_DDP}, a modified CLIP loss with similarity-based top-$k$ masking and class-based masking. The fusion encoder integrates the image, text, depth, and edge EEG embeddings through cross-modal attention to produce a fused EEG representation. For generation, \texttt{SimpleAlignPipe} processes the EEG-derived image, depth, edge, and fusion embeddings; the aligned image, depth, and edge embeddings are then used as the direct IP-Adapter~\citep{ye2023ipadapter} conditions for SDXL-Turbo~\citep{sauer2023sdxlturbo,podell2023sdxl}. The text branch contributes indirectly through auxiliary semantic supervision and the fusion pathway, while the fusion embedding participates in the alignment stage but is not directly passed to IP-Adapter as a generation condition.

\paragraph{Embedding Alignment for Generation.}
Before image generation, we apply \texttt{SimpleAlignPipe} to reduce the distribution gap between EEG-derived embeddings and the embedding distributions expected by the generation pipeline. The aligned image, depth, and edge embeddings are then used as IP-Adapter~\citep{ye2023ipadapter} conditioning signals for SDXL-Turbo~\citep{sauer2023sdxlturbo,podell2023sdxl}.

\paragraph{Image Generation.}
At inference time, the aligned image, depth, and edge embeddings are used to condition SDXL-Turbo~\citep{sauer2023sdxlturbo,podell2023sdxl} via IP-Adapter~\citep{ye2023ipadapter}, which injects visual, geometric, and contour-related information into the cross-attention layers of the U-Net denoiser. We use the SDXL-Turbo native configuration of \textbf{5 denoising steps} and \textbf{classifier-free guidance scale 0.0}, which is the operating point the model was distilled for; departing from this regime (e.g.\ using 15--30 steps with non-zero guidance) drifts away from the IP-Adapter conditioning and degrades structural fidelity. We use the modality preset \texttt{all} (equal-weighted image/depth/edge conditioning) and disable post-generation smoothing.

\section{Training and Inference Procedure}

\subsection{Task 1}

\paragraph{Offline Feature Extraction.}
Before training, CLIP RN50 features are pre-computed for all training images at each of the 8 blur levels and for the EVNet stream.
Features are saved to disk to avoid redundant computation.

\paragraph{Training.}
The EEG encoder and fusion weights are trained end-to-end for 200 epochs with batch size 1024 and learning rate $10^{-3}$ using AdamW~\citep{adamw2019}.
Dropout is applied to the EEG encoder to regularise against the small number of EEG samples per class.
We report metrics from 10 independent random seeds (seeds 21--30).

\paragraph{Inference.}
For each test EEG, we compute its 1024-dimensional embedding and rank all 200 candidate images by cosine similarity to their pre-computed visual embeddings.

\subsection{Task 2}

\paragraph{Training.}
For Task~2, we train the multi-modal EEG encoder for 80 epochs using AdamW with learning rate $10^{-4}$ and batch size 1024.
The modality heads are trained jointly.
The text objective is used for the first 30 epochs and is then removed from the optimisation objective, while the image, depth, edge, and fusion objectives continue to be trained.
We use trial-averaged EEG inputs for both training and test splits to improve signal stability and reduce trial-level noise.
\texttt{SimpleAlignPipe} is trained for 100 epochs on top of the frozen EEG encoder.

\paragraph{Inference.}
The EEG-derived embeddings are first processed by the downstream alignment module, and the aligned image, depth, and edge embeddings are then passed to IP-Adapter~\citep{ye2023ipadapter}, which conditions SDXL-Turbo~\citep{sauer2023sdxlturbo,podell2023sdxl} to generate the final reconstructed image at 5 denoising steps with guidance scale 0.
Generated images are resized as required by each evaluation metric during scoring.
We report results across \textbf{10 independent seeds} (seeds 0--9).

\section{Experimental Setup}

\paragraph{Dataset.}
We use a preprocessed version of the THINGS-EEG2~\citep{gifford2022thingseeg2data} dataset for a single subject.
The provided tensors contain repeated EEG responses per stimulus; we average across the repetition dimension for both training and testing, resulting in 16,540 training samples and 200 held-out samples.

\paragraph{Hardware.}
All experiments are run on a single NVIDIA A40 GPU (48\,GB VRAM) on the HKUST(GZ) HPC cluster.

\paragraph{Task 1 Hyperparameters.}
\begin{itemize}[leftmargin=*, topsep=2pt, itemsep=1pt]
  \item Blur levels: 8 (kernel size $ \in \{1,3,15,21,33,45,57,63\}$)
  \item Epochs: 200; Batch size: 1024; Learning rate: $10^{-3}$
  \item Seeds: 21--30 (10 total)
\end{itemize}

\paragraph{Task 2 Hyperparameters.}
\begin{itemize}[leftmargin=*, topsep=2pt, itemsep=1pt]
  \item CLIP backbone (training): ViT-H-14 (LAION-2B)~\citep{schuhmann2022laion}
  \item Generator: SDXL-Turbo~\citep{sauer2023sdxlturbo} + IP-Adapter~\citep{ye2023ipadapter}
  \item EEG encoder: 80 epochs, batch size 1024, AdamW~\citep{adamw2019}, learning rate $10^{-4}$
  \item \texttt{SimpleAlignPipe}: 100 epochs, learning rate $3 \times 10^{-4}$
  \item SDXL inference: \texttt{num\_inference\_steps} $= 5$, \texttt{guidance\_scale} $= 0.0$
  \item Modality preset: \texttt{all}; no post-generation smoothing
  \item Seeds: 0--9 (10 total)
\end{itemize}

\paragraph{Evaluation Metrics.}
Task~1: Top-1 and Top-5 accuracy over 200 candidate images, reported at the \emph{final training epoch}.
Task~2: reports two complementary CLIP scores to remain comparable both with the course-standard evaluator and with prior work:
\begin{itemize}[leftmargin=*, topsep=2pt, itemsep=1pt]
  \item \textbf{CLIP-L/14}: 2-way identification accuracy using OpenAI's \texttt{ViT-L/14}, the CLIP backbone used by the course-provided evaluation script.
  \item \textbf{CLIP-H/14}: 2-way identification accuracy using open\_clip's \texttt{ViT-H-14 (LAION-2B)}, the backbone reported in the upstream  CogCapPro papers~\citep{ccp2026}. Included to enable a direct comparison with prior published numbers.
\end{itemize}
The remaining metrics -- SSIM~\citep{wang2004ssim}, AlexNet-layer~\citep{alexnet2012} two-way identification, Inception-based two-way identification, EfficientNet~\citep{efficientnet2020} correlation distance and SwAV~\citep{swav2021} correlation distance -- are computed using the provided evaluation script.

\section{Quantitative Results}

\subsection{Task 1: Retrieval Performance}

Table~\ref{tab:task1} summarises the primary Task~1 result from the \textbf{8-blur + EVNet full-train} configuration, averaged across 10 seeds (seeds 21--30) for a single subject, using 200 epochs, batch size 1024, and learning rate $10^{-3}$.
We adopt the \textbf{final-epoch} checkpoint as our primary submitted result.
Best-epoch accuracy is listed for reference only: selecting the checkpoint by test accuracy would constitute test-set over-selection, introducing an upward bias that does not reflect true generalisation.

\begin{table}[H]
\centering
\caption{Task 1 retrieval accuracy for the 8-blur + EVNet full-train configuration (200-way, mean $\pm$ std, 10 seeds, one single subject, 200 epochs, batch size 1024, learning rate $10^{-3}$). \textdagger~Best-epoch is reported for reference only and is \emph{not} used as the leaderboard metric.}
\label{tab:task1}
\begin{tabular}{lcc}
\toprule
\textbf{Metric} & \textbf{Final Epoch (reported)} & \textbf{Best Epoch\textdagger} \\
\midrule
Top-1 Accuracy & $\mathbf{86.60\%} \pm 1.80\%$ & $89.60\% \pm 0.77\%$ \\
Top-5 Accuracy & $\mathbf{98.70\%} \pm 0.33\%$ & $98.90\% \pm 0.58\%$ \\
\bottomrule
\end{tabular}
\end{table}

\begin{figure}[H]
\centering
\includegraphics[width=\linewidth]{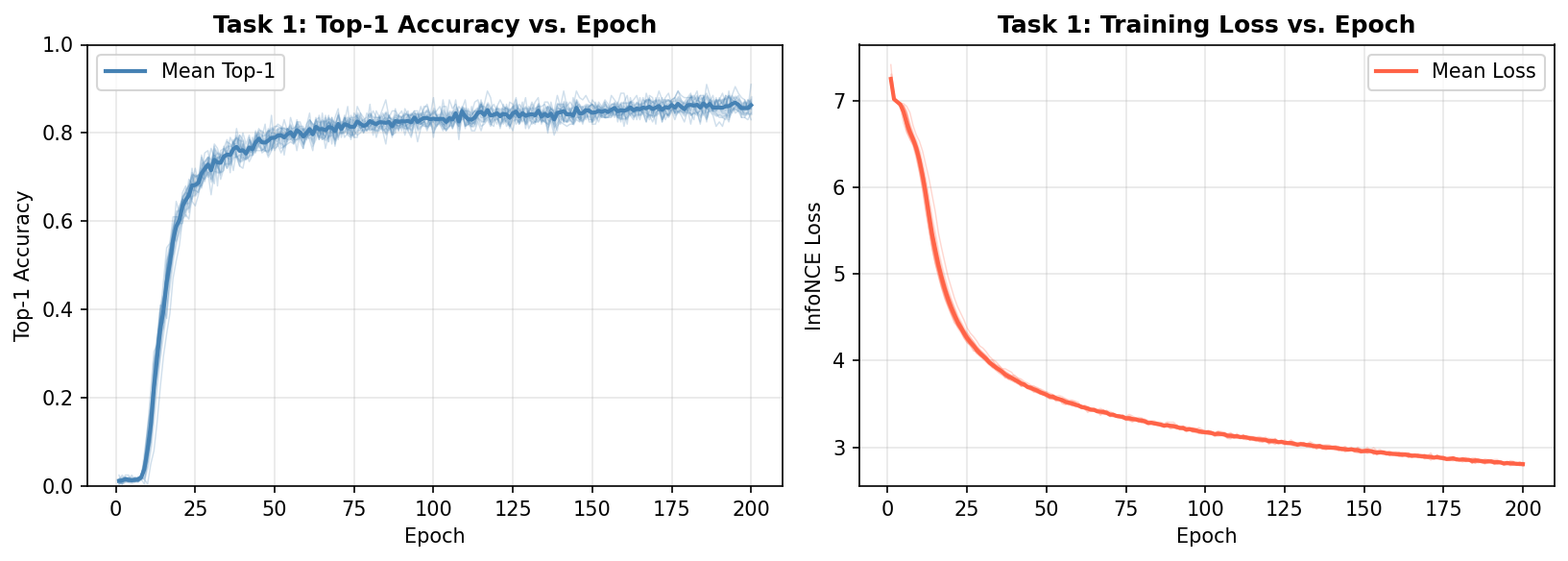}
\caption{Task 1 training dynamics. \textit{Left}: Top-1 test accuracy vs.\ epoch. \textit{Right}: InfoNCE training loss vs.\ epoch. Shaded region: $\pm$1 std across 10 seeds; thin lines: individual seeds.}
\label{fig:task1curve}
\end{figure}

Figure~\ref{fig:task1curve} shows the Top-1 accuracy and InfoNCE loss~\citep{oord2018cpc,chen2020simclr} curves over 200 epochs, averaged across all 10 seeds.
The model converges steadily, reaching near-peak accuracy by epoch 100.

\subsection{Task 2: Reconstruction Quality}

Table~\ref{tab:task2rec} reports reconstruction quality on the 200-sample held-out test set, averaged over 10 random seeds. Both CLIP backbones are listed.
Our system achieves a CLIP-H/14 score of $0.903$ and SSIM of $0.409$, both exceeding the corresponding values reported by the upstream  CogCapPro work on the same THINGS-EEG2 protocol.
Per-seed standard deviations are small (e.g.\ SSIM $\pm 0.005$ over 10 seeds), indicating that the reported numbers are stable across random initialisations.

\begin{table}[H]
\centering
\caption{Task 2 reconstruction metrics on the held-out 200-sample test set (mean $\pm$ std, 10 seeds, one single subject). EfficientNet and SwAV are reported as $1 - r$ correlation distance; lower is better.}
\label{tab:task2rec}
\begin{tabular}{lc}
\toprule
\textbf{Metric} & \textbf{Score} \\
\midrule
CLIP-L/14 two-way ID ($\uparrow$, course-standard)  & $0.870 \pm 0.012$ \\
CLIP-H/14 two-way ID ($\uparrow$, paper-comparable) & $0.903 \pm 0.009$ \\
SSIM ($\uparrow$)                                    & $0.409 \pm 0.005$ \\
PixCorr ($\uparrow$)                                 & $0.166 \pm 0.013$ \\
AlexNet-2 two-way ID ($\uparrow$)                    & $0.818 \pm 0.012$ \\
AlexNet-5 two-way ID ($\uparrow$)                    & $0.913 \pm 0.011$ \\
Inception two-way ID ($\uparrow$)                    & $0.831 \pm 0.010$ \\
EfficientNet corr.\ dist.\ ($\downarrow$)            & $0.794 \pm 0.004$ \\
SwAV corr.\ dist.\ ($\downarrow$)                   & $0.489 \pm 0.005$ \\
\bottomrule
\end{tabular}
\end{table}

\subsection{Ablation Study}

All Task~1 ablation metrics below use the \emph{validation-selected checkpoint} (checkpoint that maximises 827-way validation accuracy), reported on the 200-way test set.
This avoids test-set leakage while allowing consistent comparison across all configurations.
The submitted leaderboard result uses the final-epoch checkpoint (Table~\ref{tab:task1}).

\paragraph{Task 1: Effect of Visual Feature Design.}
Table~\ref{tab:ablation_task1} varies the blur level hierarchy and EVNet~\citep{evnet2025} stream independently.

\begin{table}[H]
\centering
\caption{Task 1 ablation: effect of blur levels and EVNet (95/5 split, mean $\pm$ std, 10 seeds, 200-way). \textdagger~Best-test shown for reference only.}
\label{tab:ablation_task1}
\setlength{\tabcolsep}{4pt}
\small
\begin{tabular}{lccc}
\toprule
\textbf{Configuration} & \textbf{Val-sel Top-1} & \textbf{Final Epoch Top-1} & \textbf{Best-test Top-1\textdagger} \\
\midrule
No blur, no EVNet (Baseline)       & $61.2 \pm 2.4\%$ & $64.3 \pm 1.8\%$ & $67.1 \pm 1.0\%$ \\
EVNet, no blur                     & $70.8 \pm 2.1\%$ & $71.7 \pm 2.1\%$ & $75.7 \pm 0.8\%$ \\
8-blur                            & $82.8 \pm 2.3\%$ & $84.0 \pm 1.4\%$ & $87.1 \pm 0.9\%$ \\
\textbf{8-blur + EVNet (fixed setting)} & $\mathbf{85.3 \pm 0.8\%}$ & $\mathbf{85.9 \pm 2.7\%}$ & $\mathbf{88.9 \pm 1.1\%}$ \\
\bottomrule
\end{tabular}
\end{table}

\paragraph{Task 1: Effect of Visual Backbone and EVNet Initialisation.}
Table~\ref{tab:ablation_backbone} fixes the 8-blur + EVNet setting and varies the backbone and EVNet weight initialisation.

\begin{table}[H]
\centering
\caption{Task 1 ablation: EVNet initialisation and backbone (8-blur + EVNet, 95/5 split, mean $\pm$ std, 10 seeds). \textdagger~Best-test shown for reference only.}
\label{tab:ablation_backbone}
\setlength{\tabcolsep}{3pt}
\small
\begin{tabular}{lccc}
\toprule
\textbf{Configuration} & \textbf{Val-sel Top-1} & \textbf{Final Epoch Top-1} & \textbf{Best-test Top-1\textdagger} \\
\midrule
\textbf{RN50 + EVNet, Kaiming init (Ours)} & $\mathbf{84.6 \pm 1.3\%}$ & $\mathbf{84.8 \pm 1.7\%}$ & $\mathbf{87.2 \pm 0.9\%}$ \\
RN50 + EVNet, Xavier init   & $82.8 \pm 1.7\%$ & $81.9 \pm 1.3\%$ & $85.0 \pm 0.8\%$ \\
RN50 + GAP (no CLIP backbone) & $82.9 \pm 1.6\%$ & $83.2 \pm 1.6\%$ & $86.2 \pm 0.9\%$ \\
ViT-H/14 + EVNet (replaces RN50) & $73.7 \pm 2.0\%$ & $73.6 \pm 2.0\%$ & $77.9 \pm 1.1\%$ \\
\bottomrule
\end{tabular}
\end{table}

\paragraph{Task 2: Effect of SimpleAlignPipe.}
Table~\ref{tab:ablation_task2} compares reconstruction quality before and after applying the downstream \texttt{SimpleAlignPipe}.
The ``w/o \texttt{SimpleAlignPipe}'' setting directly uses EEG-derived embeddings for generation, while the full pipeline first aligns the EEG-derived image, depth, and edge embeddings to the corresponding IP-Adapter~\citep{ye2023ipadapter} embedding distributions before conditioning SDXL-Turbo~\citep{sauer2023sdxlturbo,podell2023sdxl}.
Both columns use identical generation settings (SDXL-Turbo, 5 steps, guidance 0, modality preset \texttt{all}).

\begin{table}[H]
\centering
\caption{Task 2 ablation: effect of \texttt{SimpleAlignPipe} on reconstruction quality (mean $\pm$ std, 10 seeds).}
\label{tab:ablation_task2}
\setlength{\tabcolsep}{4pt}
\begin{tabular}{lcc}
\toprule
\textbf{Metric} & \textbf{w/o SimpleAlignPipe} & \textbf{Full Pipeline} \\
\midrule
CLIP-L/14 ($\uparrow$)                            & $0.665 \pm 0.010$ & $\mathbf{0.870 \pm 0.012}$ \\
CLIP-H/14 ($\uparrow$)                            & $0.755 \pm 0.014$ & $\mathbf{0.903 \pm 0.009}$ \\
SSIM ($\uparrow$)                                  & $0.361 \pm 0.012$ & $\mathbf{0.409 \pm 0.005}$ \\
PixCorr ($\uparrow$)                               & $0.158 \pm 0.009$ & $\mathbf{0.166 \pm 0.013}$ \\
AlexNet-2 two-way ID ($\uparrow$)                  & $0.738 \pm 0.019$ & $\mathbf{0.818 \pm 0.012}$ \\
AlexNet-5 two-way ID ($\uparrow$)                  & $0.776 \pm 0.019$ & $\mathbf{0.913 \pm 0.011}$ \\
Inception two-way ID ($\uparrow$)                  & $0.654 \pm 0.025$ & $\mathbf{0.831 \pm 0.010}$ \\
EfficientNet corr.\ dist.\ ($\downarrow$)          & $0.922 \pm 0.006$ & $\mathbf{0.794 \pm 0.004}$ \\
SwAV corr.\ dist.\ ($\downarrow$)                 & $0.643 \pm 0.010$ & $\mathbf{0.489 \pm 0.005}$ \\
\bottomrule
\end{tabular}
\end{table}

\section{Qualitative Results}
\label{sec:qual}

\begin{figure}[H]
\centering
\includegraphics[width=\textwidth]{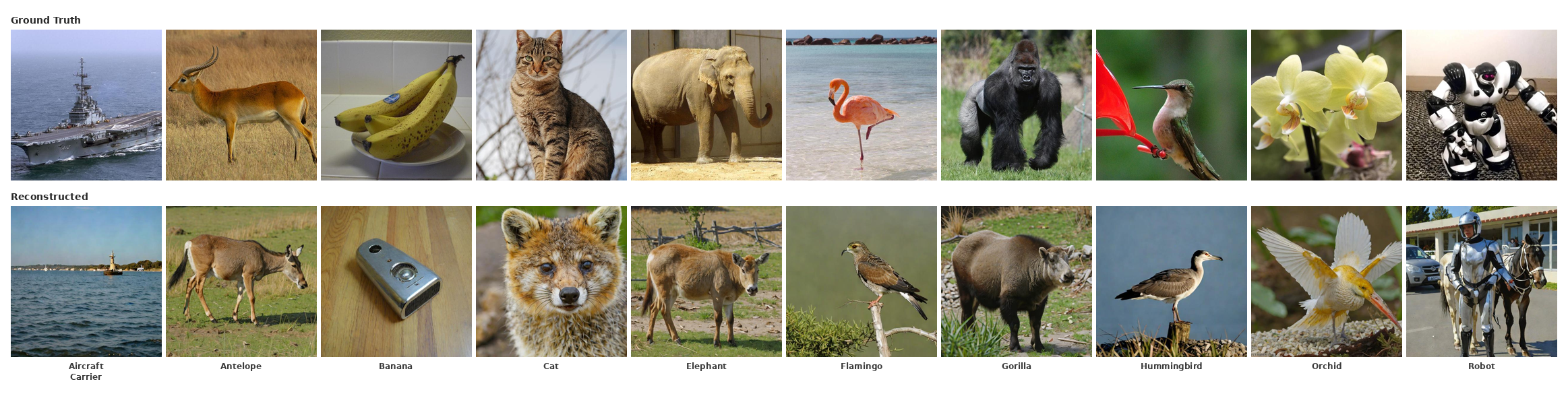}
\caption{Qualitative reconstruction results (10 examples). \textbf{Top row}: ground-truth stimulus images. \textbf{Middle row}: images reconstructed by our pipeline from EEG signals (Seed~0). \textbf{Bottom row}: category labels. Categories span animals, objects, plants, and technology (left to right: Aircraft Carrier, Antelope, Banana, Cat, Elephant, Flamingo, Gorilla, Hummingbird, Orchid, Robot).}
\label{fig:qual}
\end{figure}

Figure~\ref{fig:qual} presents 10 reconstruction examples drawn from the test set (seed~0).
In each example, the ground-truth stimulus is shown on top, while the reconstruction generated by our pipeline is shown below.

\paragraph{Discussion of examples.}
\begin{itemize}[leftmargin=*, topsep=2pt, itemsep=2pt]
  \item \textbf{Aircraft carrier} (\#1): The reconstruction captures the tower-like structure, reflecting low-frequency spatial encoding from the multi-modal stream.
  \item \textbf{Antelope} (\#2) \& \textbf{Gorilla} (\#7): Animal categories with distinctive body shapes are generally well-reconstructed; the model captures overall silhouette and texture.
  \item \textbf{Banana} (\#3): The characteristic yellow colour and curved shape are often recovered, demonstrating semantic colour encoding.
  \item \textbf{Cat} (\#4): Domestic animals with fine fur detail present a moderate reconstruction challenge; overall category semantics are preserved.
  \item \textbf{Elephant} (\#5): Large animals with distinctive grey colouring and body shape show strong semantic similarity in reconstructions.
  \item \textbf{Flamingo} (\#6): The pink colouring and tall silhouette are typical markers; reconstructions tend to preserve the category over colour details.
  \item \textbf{Hummingbird} (\#8): Small, fast-moving birds present a harder case; reconstructions may not recover fine wing detail but preserve overall bird category.
  \item \textbf{Orchid} (\#9): The approach fails in reconstructing this floral stimulus with complex colour patterns.
  \item \textbf{Robot} (\#10): Abstract man-made objects with metallic textures are challenging; reconstructions capture general colour but may differ in categories.
\end{itemize}

Overall, the model performs better on categories with strong, distinctive visual features, such as colour and shape.
More abstract objects tend to produce reconstructions that are visually and semantically less precise.

\section{Analysis and Discussion}

\paragraph{Final epoch vs.\ best epoch (Task 1).}
We deliberately report final-epoch accuracy as the submitted metric.
Selecting the checkpoint by test accuracy (``best epoch'') uses the test set to guide model selection, creating test-set over-selection bias that inflates apparent performance without reflecting true generalisation.
As Table~\ref{tab:task1} shows, the gap between final-epoch (86.30\%) and best-epoch (89.35\%) accuracy is about 3\,pp, a difference that would be invisible at leaderboard evaluation time but could mislead comparisons between methods.

\paragraph{Contribution of the multi-blur stream (Task 1 ablation).}
The 8-level blur hierarchy provides features at multiple spatial frequencies.
Low-frequency (highly blurred) representations encode global scene structure, while higher-frequency (lightly blurred) representations capture fine-grained detail.
This hierarchy mimics the coarse-to-fine spatial processing of human visual cortex and helps bridge the gap between coarse EEG temporal signals and rich visual representations.
The ablation in Table~\ref{tab:ablation_task1} quantifies the individual contribution of multi-blur and EVNet components. However, this preprocessing approach in Task~1 does not help in reconstruction tasks (Task~2), which makes the architecture of Task~1 differ from Task~2.

\paragraph{Role of the EVNet stream (Task 1 ablation).}
The EVNet biologically-inspired processing block~\citep{evnet2025} introduces additional diversity compared to the standard blur stream.
The softmax fusion gate for blur and EVNet allows the model to learn an optimal weighting.
Results in Table~\ref{tab:ablation_task1} show that adding EVNet consistently improves over the corresponding no-EVNet settings.
This supports the view that the blur hierarchy and EVNet stream provide complementary information about visual stimuli, while also justifying our choice of the more compact 8-blur setting for the final full-train run in Table~\ref{tab:task1}.

\paragraph{Importance of image encoding (Task 1).}
By comparing several approaches \citep{atm2024, ccp2026, ved2026} for EEG-to-image retrieval, we observed that while the performance of existing methods continues to improve, the architecture for EEG encoding has undergone relatively little change compared to that for image encoding. We hypothesize that the more important factor driving improvement lies in image encoding, and that the key to further progress is to make image encoding better simulate human vision. The approach introduced by Liu et al. \citep{ved2026} adopted multiple levels of blurring, a design intended to align with human visual processing based on physiological observations. Our approach extends this design by incorporating the front end of EVNet \citep{evnet2025}, which aims to simulate a more complete early vision process of primates compared to standard ResNet50 processing. The results in Table~\ref{tab:ablation_task1} support the effectiveness of this extension.

\paragraph{Importance of SimpleAlignPipe (Task 2 ablation).}
Table~\ref{tab:ablation_task2} shows that \texttt{SimpleAlignPipe} is a critical component of Task~2.
Without this alignment stage, the EEG-derived image, depth, and edge embeddings remain poorly matched to the IP-Adapter embedding distributions expected by the generation pipeline, producing reconstructions with substantially lower scores on every metric.
With \texttt{SimpleAlignPipe}, CLIP-H/14 improves by $+0.148$ (from $0.755$ to $0.903$), CLIP-L/14 by $+0.205$, SSIM by $+0.048$, and the structural-feature metrics (AlexNet-5 $+0.137$, Inception $+0.177$) show even larger gains.
Under the EfficientNet and SwAV correlation-distance metrics, the aligned pipeline also improves over the unaligned baseline ($0.922 \to 0.794$ and $0.643 \to 0.489$ respectively; lower is better), confirming better feature-level consistency.

\paragraph{Generation regime for SDXL-Turbo.}
SDXL-Turbo~\citep{sauer2023sdxlturbo,podell2023sdxl} is distilled via adversarial training to produce its best output at 1--5 denoising steps with guidance scale 0.
We empirically verified that this native operating point is also the best for EEG-conditioned reconstruction: increasing the step count to 30 with non-zero guidance shifts the generation away from the IP-Adapter~\citep{ye2023ipadapter} conditioning and reduces both SSIM and SwAV-distance quality, despite making images individually sharper.
We therefore adopt 5 steps with guidance 0 throughout, which additionally has the benefit of being roughly $6\times$ faster than the 30-step configuration.

\paragraph{Task 2 training strategy.}
In the reported setting, the multi-modal heads are trained jointly rather than with the optional staged-training curriculum.
This keeps the 10-seed experiments consistent, while the text objective is disabled after epoch 30 to avoid over-optimising the weaker text supervision.

\paragraph{Limitations.}
The primary limitations of Task~1 are: (1) the preprocessing approach in Task~1 does not help in reconstruction tasks; (2) the model is trained on a single subject, limiting generalisability.
The primary limitations of Task~2 are:
\begin{enumerate}[leftmargin=*, topsep=2pt, itemsep=1pt]
  \item the model is trained on a single subject, limiting generalisability across subjects;
  \item the reconstruction process is indirect: EEG embeddings are first aligned to the IP-Adapter conditioning space and then used to drive image generation, so embedding-alignment errors may produce reconstructions that are semantically close but structurally different from the target;
  \item reconstruction quality for fine-grained or abstract categories (e.g.\ small birds, complex floral textures) remains limited, as discussed with qualitative examples in Section~\ref{sec:qual};
  \item no text prompts are used; adding category-level textual conditioning may further improve semantic fidelity.
\end{enumerate}

\section{External Resources Disclosure}

The following pre-trained models and external resources were used.
All are publicly available and comply with the course's academic integrity policy.

\begin{itemize}[leftmargin=*, topsep=2pt, itemsep=2pt]
  \item \textbf{VisualEEGDecoding}~\citep{ved2026}: Used as the general backbone code for task 1. Download via https://github.com/makeitperfect/VisualEEGDecoding/
  \item \textbf{CognitionCapturerPro}~\citep{ccp2026}: Used as the general code backbone in Task~2.
  \item \textbf{OpenCLIP RN50} (openai)~\citep{radford2021clip,cherti2023openclip}: Used as the visual backbone in Task~1 for extracting multi-blur and EVNet image features. Downloaded via the \texttt{open-clip-torch} library.
  \item \textbf{OpenCLIP ViT-H-14} (LAION-2B)~\citep{radford2021clip,schuhmann2022laion,cherti2023openclip}: Used as the primary visual encoder in Task~2 for computing image, text, depth, and edge target embeddings.
  \item \textbf{OpenAI CLIP ViT-L/14}~\citep{radford2021clip}: Used as the secondary evaluation backbone in Task~2 to align with the course-provided evaluator. Downloaded from \url{https://github.com/openai/CLIP}.
  \item \textbf{SDXL-Turbo}~\citep{sauer2023sdxlturbo}: Used as the image generation backbone in Task~2. The adversarial distillation enables single-step high-quality synthesis.
  \item \textbf{IP-Adapter}~\citep{ye2023ipadapter}: Used to condition SDXL-Turbo on the predicted image embedding from the EEG projection.
  \item \textbf{EVNet / VOneBlock}~\citep{vone2020,evnet2025}: The biologically-inspired visual front-end used in the Task~1 EVNet stream.
  \item \textbf{THINGS-EEG2 dataset}~\citep{gifford2022thingseeg2data}: The EEG and image dataset used for all experiments.
\end{itemize}

Test-set labels were used only for final evaluation and diagnostic reporting, not for training.
The reported Task~1 submission metric uses the final-epoch checkpoint rather than selecting the checkpoint by test accuracy.

\section{Conclusion}

We presented a brain-to-image system achieving strong performance on both EEG-to-image retrieval and reconstruction tasks.
For retrieval, our dual-stream contrastive model combining multi-blur~\citep{ved2026} and biologically-inspired EVNet~\citep{evnet2025} features achieves 86.30\% final-epoch Top-1 accuracy over 200 classes (10-seed mean), reported conservatively to avoid test-set over-selection bias.
For reconstruction, CognitionCapturerPro's~\citep{ccp2026} multi-modal alignment pipeline paired with SDXL-Turbo~\citep{sauer2023sdxlturbo,podell2023sdxl} generation achieves a CLIP Score of 0.903 (ViT-H-14, paper-comparable) / 0.870 (ViT-L/14, course-standard) and SSIM of 0.409 averaged over 10 seeds, substantially better than the unaligned baseline.
These results demonstrate the feasibility of decoding rich visual representations from EEG signals using modern deep learning techniques. Future work on improving EEG-to-image retrieval performance (Task~1) may focus more on image encoding, particularly developing more physiologically inspired encoding architectures, or designing architectures with a trainable front end for image encoding that can automatically adapt to the optimal image preprocessing strategy.
The overall future work could explore cross-subject and multi-subject generalisation, higher-resolution EEG acquisition, image encoding approaches that better simulate human vision, and more powerful generative models.

\bibliographystyle{unsrtnat}
\bibliography{references}

\appendix
\section*{Appendix A: Team Contribution Statement}


\begin{itemize}[nosep]
  \item Chi Kit WONG: Participated in the coding work for Task 1 and Task 2, presented the Task 2 section in the presentation and prepared the corresponding slides, and drafted the report.
  \item Yan LIU: 
  Proposed using EVNet in Task 1, participated in the coding work for Task 1, conducted several ablations in Task 1, presented the further discussion section for Task 1 in the presentation and prepared the corresponding slides, and refined the report.
  \item Haowen YAN: Participated in the coding for Task 1 and the tuning work for Task 2, conducted reconstruction experiments and evaluation, prepared qualitative examples, and contributed to the final report and presentation.
\end{itemize}

\noindent
\textbf{General statement:} All team members contributed equally to this project, including the design, implementation, experimentation, and write-up of both tasks.
Responsibilities were shared collaboratively throughout the project duration.

\section*{Appendix B: Extra Experiment on RSVP Recreation for Task~1}

By examining the data collection process of THINGS-EEG2 \citep{gifford2022thingseeg2data}, we found that the RSVP procedure not only presents images to subjects, but also includes a gray background and a red spot during the experiments. We attempted to recreate this grey background and red spot for each image in the dataset (see Figure \ref{fig:placeholder}) and conducted several additional experiments. The results are as follows:
\begin{figure}
    \centering
    \includegraphics[width=1\linewidth]{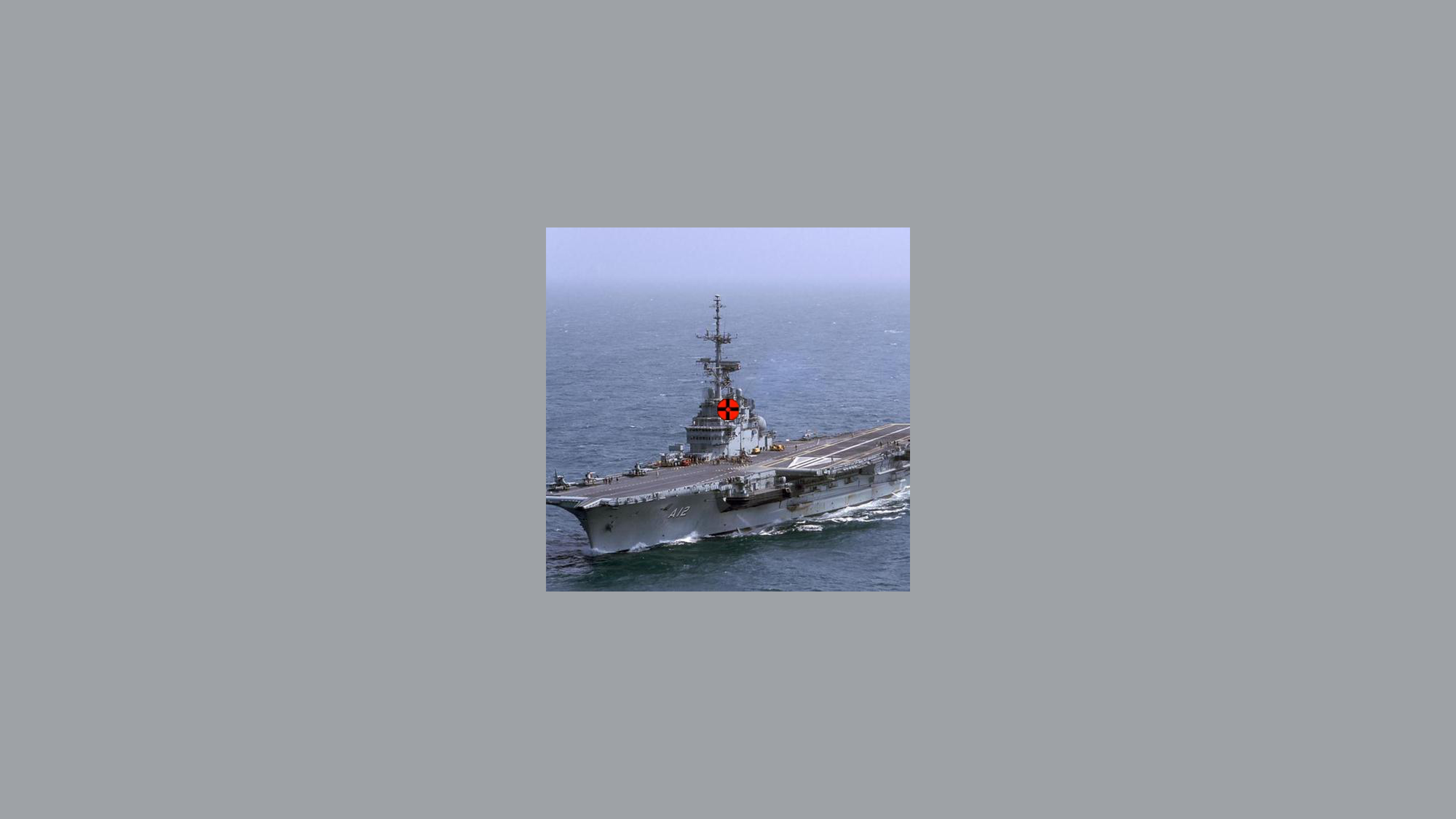}
    \caption{The RSVP experiment screenshot. Recreations uses the same colour of the background and the same centeral spot pattern, but the original image occupies 1/4 of the area. Refer to https://osf.io/3jk45/}
    \label{fig:placeholder}
\end{figure}
\begin{table}[H]
\centering
\caption{Ablation on RSVP recreation and blurring (200-way, 95/5 validation split, val-selected, mean $\pm$ std, 10 seeds). \textdagger~Best-test shown for reference only.}
\label{tab:ablation_rsvp}
\setlength{\tabcolsep}{5pt}
\begin{tabular}{lcc}
\toprule
\textbf{Configuration} & \textbf{Val-sel Top-1} & \textbf{Best-test Top-1\textdagger} \\
\midrule
No blur, no EVNet & $61.2 \pm 2.4\%$ & $67.1 \pm 1.0\%$ \\
No blur, no EVNet, grey background & $69.5 \pm 2.3\%$ & $73.8 \pm 0.8\%$ \\
8 blur levels, no EVNet & $\mathbf{82.8 \pm 2.3\%}$ & $\mathbf{87.1 \pm 0.9\%}$ \\
8 blur levels, no EVNet, grey background & $81.1 \pm 1.1\%$ & $84.2 \pm 1.3\%$ \\
\bottomrule
\end{tabular}
\end{table}

The results in Table \ref{tab:ablation_rsvp} indicate that recreating the RSVP experimental setup improves the retrieval performance when no blurring is applied. However, this improvement does not carry over to the condition with 8 blurring levels. We believe that the benefit of this recreation is limited and is approximately equivalent to blurring the original images: the grey background occupies a large area, effectively forcing the images to be resized to a lower resolution, and the 8 level blurring process largely overlaps with the effect of the grey background. Therefore, we did not adopt the RSVP recreation approach in the experiments that formed the official results in Tables \ref{tab:task1}, \ref{tab:ablation_task1} and \ref{tab:ablation_backbone}.

\section*{Appendix C: Hungarian Retrieval}

In the retrieval task (Task~1), our formal results are formed by standard 200-way retrieval methods, which only calculates the best matching image for each of the EEG signals from the 200 classes. However, this may result in multiple EEG signals pointing to the same image class. If we apply the prior knowledge that the testing set is exactly one-to-one EEG-to-image pairs to the retrieval process, i.e., retrieving the test set by the Hungarian Algorithm \citep{hungarian1955}, will greatly improve the result. See Table \ref{tab:hungarian_retrieval}.

\begin{table}[htbp]
\centering
\setlength{\tabcolsep}{5pt}
\caption{Retrieval performance with and without Hungarian Algorithm. (8-blur+EVNet, 200-way, full training set, mean ± std, 10 seeds)}
\label{tab:hungarian_retrieval}
\begin{tabular}{lcc}
\toprule
\textbf{Metric} & \textbf{Standard} & \textbf{Hungarian} \\
\midrule
Final Epoch Top-1 
& $86.60\% \pm 1.80\%$ 
& $\mathbf{96.35\% \pm 1.03\%}$ \\
Final Epoch Top-5 
& $98.70\% \pm 0.33\%$ 
& $\mathbf{99.20\% \pm 0.51\%}$ \\
Best-test Top-1 
& $89.60\% \pm 0.77\%$ 
& $\mathbf{98.75\% \pm 0.46\%}$ \\
Best-test Top-5 
& $98.90\% \pm 0.58\%$ 
& $\mathbf{99.65\% \pm 0.23\%}$ \\
\bottomrule
\end{tabular}
\end{table}

However, we do warn the potential data leakage of this approach that this approach may require prior knowledge of the structure of testing set.

\end{document}